\documentclass[10pt,twocolumn,letterpaper]{article}

\usepackage[pagenumbers]{cvpr}

\usepackage{multirow}
\newcommand{\airvt}{AirVT-15k\xspace}
\newcommand{\airfocus}{AirFocus-CST\xspace}

\DeclareMathOperator*{\argmax}{arg\,max}

\definecolor{cvprblue}{rgb}{0.21,0.49,0.74}
\usepackage[pagebackref,breaklinks,colorlinks,allcolors=cvprblue]{hyperref}

\title{AirVT-15k: A Query-Diverse UAV Video--Text Dataset and Query-Adaptive Evidence Alignment Network for Retrieval}

\author{%
\begin{tabular}{ccc}
Xi Chen & Xu Chen & Xiangyang Jia \\
{\small\ttfamily chenxi00233@whu.edu.cn} &
{\small\ttfamily xuchen@whu.edu.cn} &
{\small\ttfamily jxy@whu.edu.cn} \\[0.6em]
Ting Gan & Xu Zhang & Shuquan Wei \\
{\small\ttfamily ganting@whu.edu.cn} &
{\small\ttfamily zhangx0802@whu.edu.cn} &
{\small\ttfamily 2021302111045@whu.edu.cn}
\end{tabular}%
}

\begin{document}
\maketitle

\begin{abstract}
Unmanned aerial vehicle video--text retrieval (UAV-TVR) requires matching language to both scene context and subtle object changes. A harbor-layout query relies on global scene context, whereas a departing-vessel query requires local functional cues and motion states. Existing benchmarks provide diverse aerial descriptions but give limited attention to distinct evidence needs within the same video. We introduce \airvt, a benchmark of 15,504 UAV videos and 155,040 captions, with ten complementary bilingual captions per video describing scenes, object states, and temporal changes. We also propose \airfocus, which selects query-relevant frames and patches, encodes temporal feature differences, and combines coarse-to-fine interaction (CFI) with optimal-transport alignment (OTA). Average mean recall reaches 63.67\% on \airvt and 24.41\% on CapERA, improving over the strongest baselines by 2.96 and 1.91 percentage points, respectively.
The dataset and code are available at \url{https://chenxi000233.github.io/AirVT-15k-site/}.

\end{abstract}

\section{Introduction}
\label{sec:intro}

\begin{figure}[t]
    \centering
    \includegraphics[width=\linewidth]{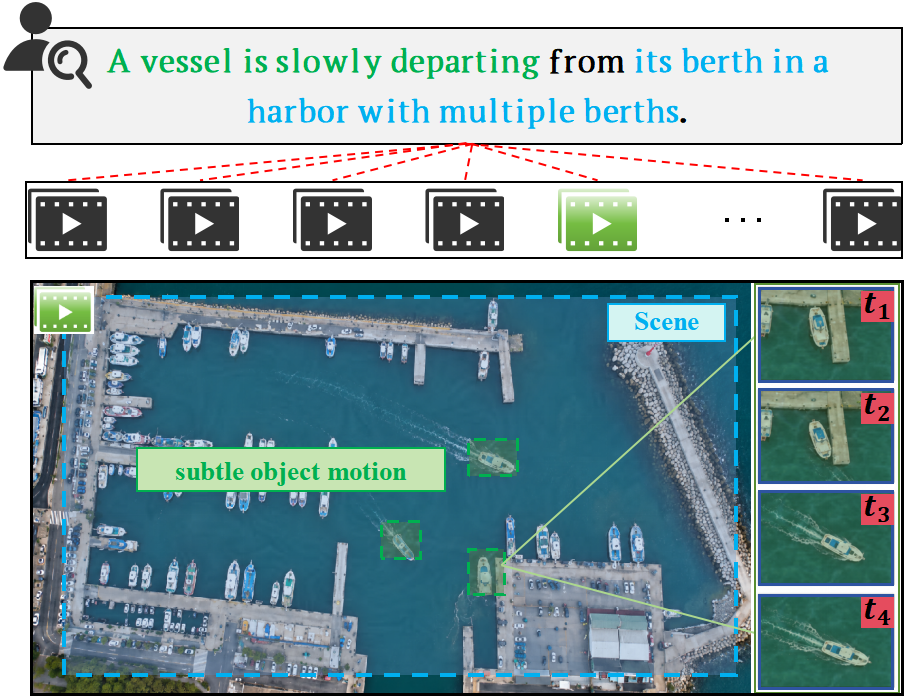}
    \caption{Query-dependent visual evidence in UAV video--text retrieval.
    Descriptions of scene context and object changes within the same video
    require different combinations of global scene context, local functional
    cues, and motion states.}
    \label{fig:teaser}
\end{figure}

UAV video--text retrieval supports traffic inspection, port monitoring, and disaster assessment. Queries about the same video can require different visual evidence. A harbor-layout query refers to docks and waterways; a departing-vessel query refers to the vessel's position relative to its berth and how that position changes over time (Fig.~\ref{fig:teaser}). We describe these evidence types as global scene context, local functional cues, and motion states. Wide views, small objects, and subtle motion make their joint representation difficult.

General video--text benchmarks such as MSR-VTT~\cite{xu2016msrvtt} cover diverse content, and aerial resources such as GeoChat~\cite{kuckreja2024geochat} support image-level understanding. UAV video benchmarks, including DVTMD~\cite{zhao2025tcma}, extend this setting to aerial video descriptions. Evaluating query-dependent evidence needs also requires captions that describe different aspects of the same clip. We introduce \airvt with 15,504 videos covering aerial scenes, object activities, and temporal changes. Each video has ten complementary bilingual captions, so queries can refer to its surroundings, local objects, or changes over time. Video-level data splits and bidirectional retrieval protocols define the training and evaluation setting.

Retrieval methods address frame aggregation and cross-modal correspondence in different ways. CLIP4Clip~\cite{luo2021clip4clip} aggregates frame representations, X-Pool~\cite{gorti2022xpool} pools frames conditioned on text, and TCMA~\cite{zhao2025tcma} aligns sentences with frames and words with patches. UAV retrieval also requires distinguishing activities among videos with similar scenes and object appearances. We propose \airfocus with two components: Query-Guided Multi-Evidence Modeling and Complementary Fine-Grained Cross-Modal Alignment. The first retains global scene context and combines query-guided temporal selection (TS), query-guided spatial refinement (SR), and temporal variation modeling (TV). TS and SR focus on relevant moments and local functional cues; TV encodes frame-feature changes associated with motion states. The second combines Coarse-to-Fine Interaction (CFI) on query-refined features with Optimal-Transport Alignment (OTA) on query-independent visual features. CFI checks sentence--frame, word--frame, and word--patch matches, while OTA jointly allocates word--patch correspondence weights across the video.

Our contributions are threefold:
\begin{itemize}[leftmargin=*,nosep]
    \item We introduce \airvt, a UAV video--text benchmark with 15,504 videos, ten complementary bilingual captions per video, and fixed bidirectional retrieval protocols. The captions link each clip to queries about its scene, objects, and temporal changes.
    \item We propose \airfocus to select query-relevant visual evidence, encode temporal variation, and combine CFI with OTA for fine-grained video--text matching.
    \item Evaluation on \airvt and CapERA shows that \airfocus achieves the highest average mean recall among the compared methods. Component ablations quantify the gains from temporal selection, spatial refinement, temporal variation, and the two alignment modules.
\end{itemize}

\section{Related Work}
\label{sec:related}

\subsection{General Video--Text Benchmarks}
\label{sec:related_benchmarks}
MSR-VTT~\cite{xu2016msrvtt} pairs web videos with captions, and ActivityNet Captions~\cite{krishna2017dense} associates descriptions with temporal segments. HowTo100M~\cite{miech2019howto100m} and MIL-NCE~\cite{miech2020milnce} extend video--text learning to large collections of narrated instructional videos. Recent data-centric approaches refine noisy video annotations and expand event-level descriptions to improve the quality and coverage of video--text supervision~\cite{wang2025dataflywheel,bai2025asymmetry}. Feint6K~\cite{ma2024feint6k}, MotionBench~\cite{hong2025motionbench}, and VideoComp~\cite{kim2025videocomp} examine cross-frame reasoning, motion understanding, and compositional temporal alignment, respectively. For UAV-TVR, we pair aerial environments with captions that distinguish scene descriptions from object activities within the same clip.

\subsection{Aerial and UAV Vision--Language Benchmarks}
\label{sec:related_aerial}
GeoChat~\cite{kuckreja2024geochat} and AerialVG~\cite{liu2025aerialvg} address aerial image understanding and visual grounding. For aerial videos, CapERA~\cite{bashmal2023capera} expands human captions through paraphrasing and back-translation. DVTMD~\cite{zhao2025tcma} provides five complementary descriptions per video from the ERA source, and USRD and UMCRD~\cite{huang2026msam} support drone video--text retrieval. \airvt extends this setting with videos spanning scenes, applications, and events, paired with ten captions describing different aspects of each clip. Section~\ref{sec:dataset} describes the collection, annotation, and evaluation protocols.

\subsection{Video--Text Retrieval}
\label{sec:related_retrieval}
Video--language pretraining learns joint representations through temporal and multimodal encoding, global--local alignment, and image--video training~\cite{gabeur2020mmt,wang2021t2vlad,bain2021frozen}. Contrastive pretraining and adaptation of image--text models extend these representations to video--language tasks~\cite{xu2021videoclip,wang2023allinone,xue2023clipvip}.

Building on Contrastive Language--Image Pre-training (CLIP)~\cite{radford2021clip}, retrieval models combine frame aggregation, multi-grained correspondence, and text-conditioned pooling~\cite{luo2021clip4clip,ma2022xclip,gorti2022xpool}. Hierarchical, token-level, and intra-modality contrastive objectives improve cross-modal alignment~\cite{liu2021hit,yang2021taco,zolfaghari2021crossclr}. Token clustering and spatiotemporal token selection reduce visual redundancy~\cite{zhao2022centerclip,liu2022ts2net}. Local association and coarse-to-fine alignment connect representations at different granularities~\cite{ge2022bridgeformer,wang2023ucoa}, while distillation transfers matching knowledge to retrieval encoders~\cite{croitoru2021teachtext,gao2021clip2tv}.

Recent fine-grained models emphasize less salient discriminative objects, global--local interactions, token-wise matching, and context-aware text--frame relationships~\cite{zheng2025lsdo,bian2026davrl,reddy2025videocolbert,chen2026eaglenet}.

Recent retrieval models explore pseudo-query interaction, bidirectional likelihood estimation, and generative--contrastive training~\cite{lan2025hybridtower,ko2025blim,gupta2026ville}. Auxiliary captions, scene elements, and speech provide additional semantic cues~\cite{jang2026corecaption,le2025bima,zhao2026save}; drone-specific methods use text-conditioned alignment and adaptive semantic mining~\cite{zhao2025tcma,huang2026msam}.

Partially relevant retrieval further matches queries to video moments through multi-scale similarities and diffusion-guided contextual registers~\cite{chen2026prvr,li2026imagine}.

\paragraph{Temporal difference modeling.}
For action recognition, dynamic images~\cite{bilen2016dynamicimage} summarize temporal evolution by rank pooling, while Representation Flow~\cite{piergiovanni2019representationflow} and MotionSqueeze~\cite{kwon2020motionsqueeze} learn flow or inter-frame correspondence. STM~\cite{jiang2019stm} and TEA~\cite{li2020tea} encode feature-level motion, and TDN~\cite{wang2021tdn} models short- and long-range temporal differences. For video--text retrieval, CLIP2Video~\cite{fang2021clip2video} introduces a temporal difference block. \airfocus combines differences of frozen frame features with query-refined local evidence, then computes correspondence on visual features before and after query refinement.

\section{The AirVT-15k Benchmark}
\label{sec:dataset}
\begin{figure}[t]
    \centering
    \includegraphics[width=\linewidth]{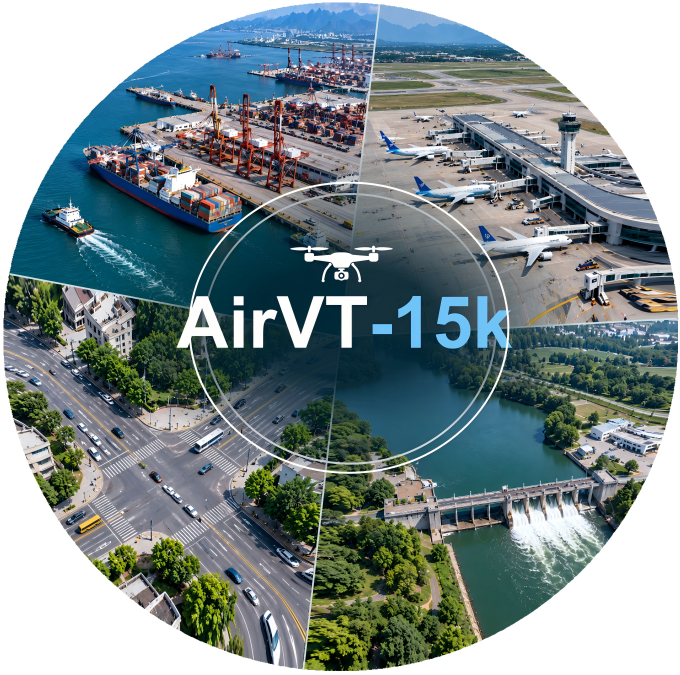}
    \caption{An overview of representative UAV scenes covered by \airvt,
    including ports, airports, urban roads, and waterways. These scenes
    provide contexts for diverse applications and events.}
    \label{fig:airvt_overview}
\end{figure}

\airvt contains UAV videos of ports, airports, urban roads, waterways, and other aerial environments (Fig.~\ref{fig:airvt_overview}). The collection covers traffic monitoring, infrastructure inspection, and area observation, with events such as moving, entering and leaving, crossing, and docking. Complementary captions describe the surroundings and the objects' activities and states, allowing the same clip to answer different queries.

\subsection{Data Collection and Curation}
\label{sec:dataset_collection}
We collect videos from our UAV recordings and public online sources.
We record footage with a DJI Mavic 3 equipped with a three-axis mechanically
stabilized camera, primarily at
$3840\times2160$ and 30 fps with H.265 encoding. Flight altitudes range from
approximately 20--120 m, with most footage captured at 30--80 m. Nadir, oblique,
tracking, and orbiting views cover roads, built-up areas, farmland,
rivers, and construction sites. Public videos from video-sharing websites, open
datasets, and project pages extend geographic, viewpoint, and semantic
coverage. Long recordings are segmented using shot boundaries and semantic
continuity into approximately 15--30 s clips, with temporal overlap when
needed to preserve complete events. We remove corrupted, low-resolution,
severely blurred or shaky clips, prolonged static footage, and clips dominated
by subtitles or editing effects. Deduplication combines file information and
keyframe perceptual hashes for coarse screening, sampled-frame visual
similarity for finer comparison, and manual review to remove duplicate clips.

\subsection{Annotation}
\label{sec:dataset_construction}
For each video, we randomly sample 8--32 frames according to its duration and
present them chronologically at their original resolution.
ChatGPT-6 generates ten English--Chinese caption pairs using a structured prompt.
The caption templates guide descriptions toward complementary aspects
of the visible scene and object activity. Each template applies to content
present in the sampled frames; additional captions describe other observable
details. Caption pairs must preserve bilingual semantic consistency and
ground objects, actions, directions, counts, and causal statements in visual evidence.

Automated checks flag event-level semantic repetition among captions. Flagged
samples undergo manual review, and captions are regenerated when problems
are identified according to the annotation quality-control rules.

\subsection{Dataset Statistics and Analysis}
\label{sec:dataset_statistics}
\begin{figure*}[t]
    \centering
    \includegraphics[width=\textwidth]{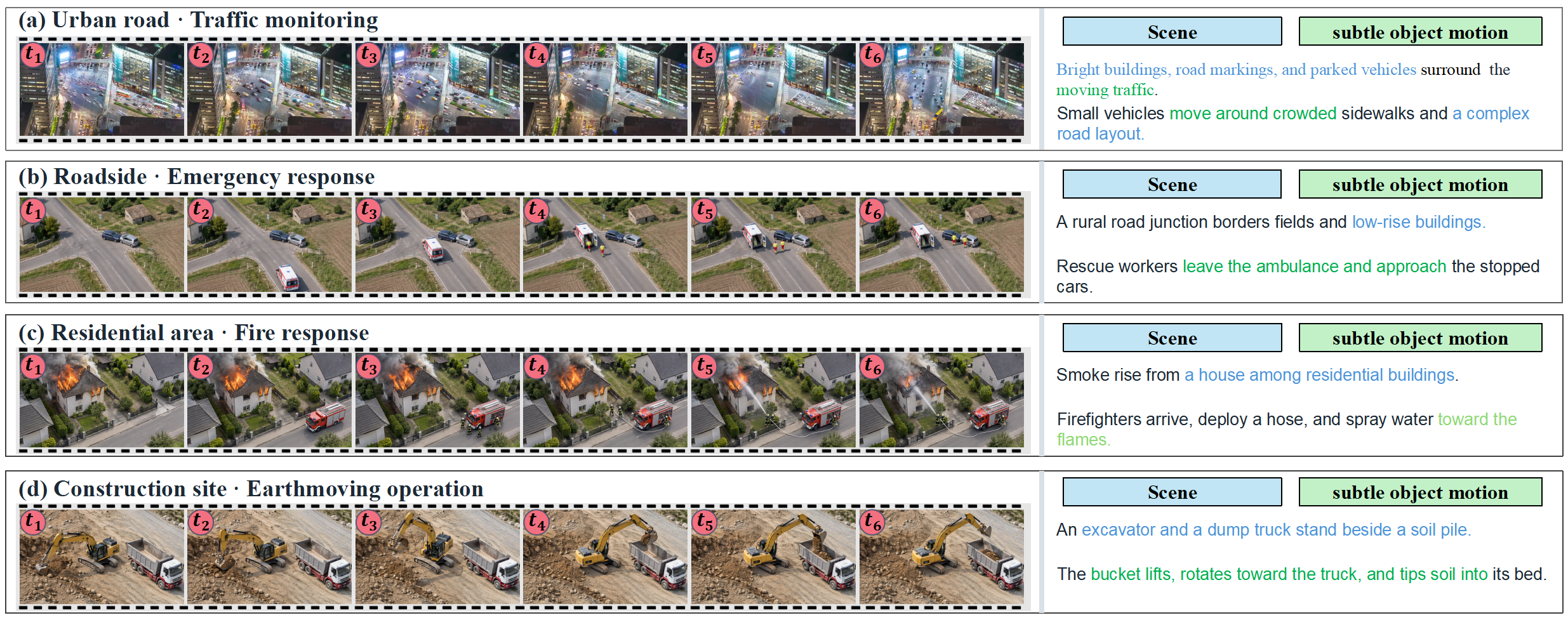}
    \caption{Chronological frames and complementary descriptions highlighting scene content, application context, and visible events within each sequence.}
    \label{fig:caption_complementarity}
\end{figure*}

Figure~\ref{fig:caption_complementarity} shows how captions describe different content within a clip. Scene descriptions identify the surrounding environment, while activity descriptions follow object states and changes across chronological frames. The captions pair the scene context needed to interpret an activity with the changes that distinguish it.

\begin{figure*}[t]
    \centering
    \includegraphics[width=\textwidth]{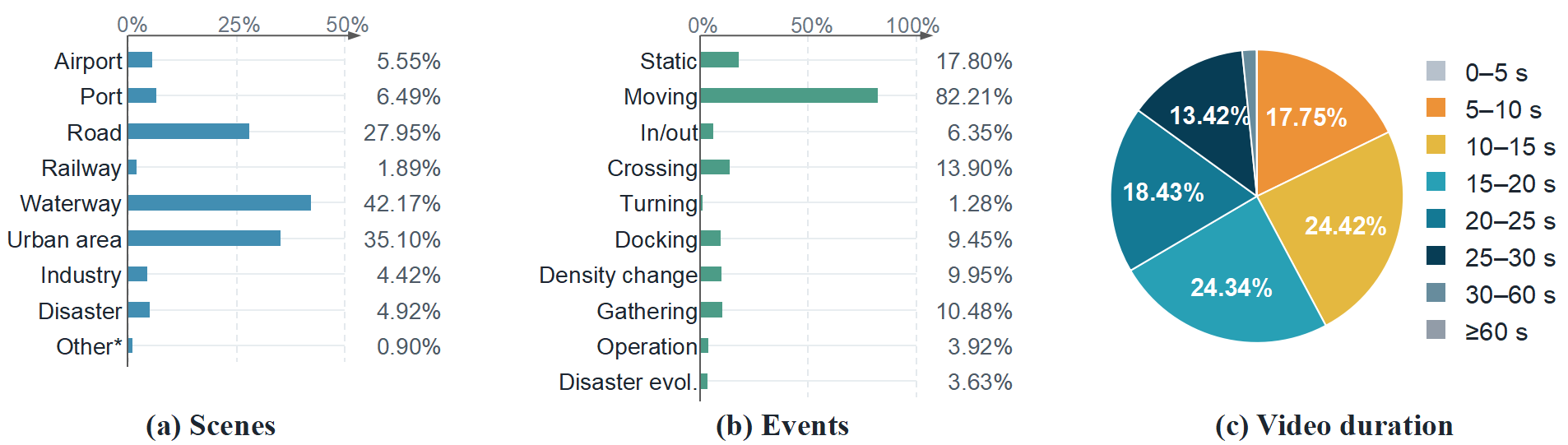}
    \caption{AirVT-15k distributions of scenes, events, and video duration. A video may belong to multiple categories (a,b).}
    \label{fig:dataset_distributions}
\end{figure*}

Figure~\ref{fig:dataset_distributions}(a,b) reports scene and event distributions derived from the captions. Waterways, urban areas, and roads are the largest scene categories, and moving objects dominate the event distribution. Entering and leaving, crossing, and docking describe more specific activities. The distributions are multi-label: a clip can belong to several scene and event categories. Figure~\ref{fig:dataset_distributions}(c) reports the distribution of video duration.

\subsection{Benchmark Tasks and Evaluation Protocols}
\label{sec:dataset_protocol}
We evaluate text-to-video and video-to-text retrieval with fixed data splits,
positive video--text pairs, and recall metrics.

\noindent\textbf{Data splits.}
\airvt contains 15,504 videos, each paired with ten complementary descriptions.
We partition the dataset into training, validation, and test subsets at an
8:1:1 ratio. All descriptions associated with a video remain in the same subset.

\noindent\textbf{Retrieval tasks.}
Text-to-video (T2V) retrieval ranks candidate videos for each description,
with its paired video as the positive. Video-to-text (V2T) retrieval ranks
candidate descriptions for each video, treating its ten paired descriptions
as positives. Candidates include all videos or all descriptions in the
corresponding evaluation subset.

\noindent\textbf{Evaluation metrics.}
We report $\mathrm{R@1}$, $\mathrm{R@5}$, and $\mathrm{R@10}$ in both directions.
For direction $d\in\{\mathrm{T2V},\mathrm{V2T}\}$, recall at $K$ (R@K) is
\begin{equation}
    \mathrm{R@K}=\frac{100}{N_d}\sum_{i=1}^{N_d}
    \mathbf{1}\!\left[r_d(q_i)\leq K\right],
    \label{eq:retrieval_recall}
\end{equation}
where $N_d$ is the number of queries, $r_d(q_i)$ is the rank of the
highest-ranked positive for query $q_i$, and $\mathbf{1}[\cdot]$ is the
indicator function. Thus, a query succeeds if at least one positive appears
in the top-$K$ results. Directional mean recall (mR) averages R@1, R@5, and R@10.
Average mean recall (Avg.\ mR) averages all six recalls:
\begin{equation}
    \mathrm{Avg.\ mR}=\frac{1}{6}\sum_{K\in\{1,5,10\}}
    \left(\mathrm{R@K}_{\mathrm{T2V}}+\mathrm{R@K}_{\mathrm{V2T}}\right).
    \label{eq:retrieval_mr}
\end{equation}

\section{AirFocus-CST}
\label{sec:method}

\begin{figure*}[t]
    \centering
    \includegraphics[width=\textwidth]{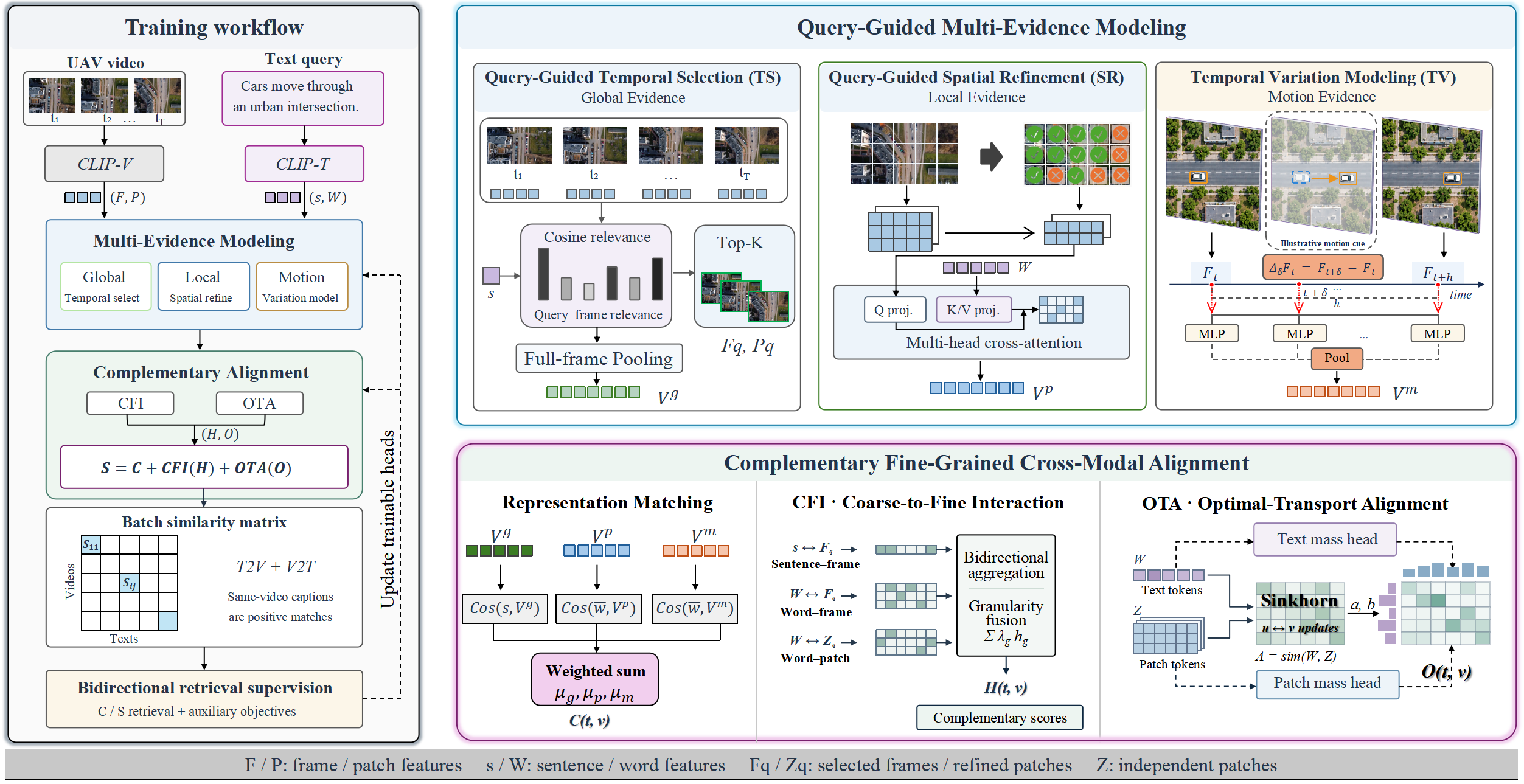}
    \caption{Overview of \airfocus. Query-guided multi-evidence modeling combines global, local, and motion information, while complementary fine-grained cross-modal alignment integrates CFI and OTA into a unified matching score.}
    \label{fig:method}
\end{figure*}

\subsection{Method Overview}
\label{sec:method_problem}
\airfocus has two components: Query-Guided Multi-Evidence Modeling and Complementary Fine-Grained Cross-Modal Alignment (Fig.~\ref{fig:method}). Given a UAV video $v$ and a text query $t$, frozen CLIP encoders extract frame features $F$, patch features $P$, a sentence feature $s$, and text token features $W$.

The evidence component retains global scene context in $V^g$, uses TS and SR to form a local representation $V^p$ of query-relevant objects and their context, and uses TV to encode temporal changes in $V^m$. These representations supply the global, local, and motion information required by different queries. The alignment component computes CFI on query-refined features and OTA on query-independent visual features. Their scores $H(t,v)$ and $O(t,v)$ are added to the representation similarity $C(t,v)$ to obtain the final score $S(t,v)$. 

\subsection{Query-Guided Multi-Evidence Modeling}
\label{sec:method_evidence}
\noindent\textbf{Query-Guided Temporal Selection (TS).}
TS selects the frames most relevant to the query and retains a full-video
representation of the surrounding scene.
Given sentence feature $s$ and frame features $F=\{F_i\}_{i=1}^{T}$, we compute
\begin{equation}
    \begin{aligned}
    r_i&=\operatorname{cos}(s,F_i),\quad
    \mathcal I_q=\operatorname{TopK}_{i}(r_i),\\
    V^g&=\frac{1}{T}\sum_{i=1}^{T}F_i.
    \end{aligned}
\end{equation}
Here, $\mathcal I_q$ contains the indices of the $K$ selected frames.
Their frame features $F_q=\{F_i\}_{i\in\mathcal I_q}$ and patch features $P_q=\{P_i\}_{i\in\mathcal I_q}$ support subsequent local modeling and alignment, while $V^g$ preserves global scene context.
Frame selection determines the inputs to SR and CFI. The global representation
still averages all sampled frames, including those outside $\mathcal I_q$,
so local matching can focus on the query without discarding the surrounding scene.

\noindent\textbf{Query-Guided Spatial Refinement (SR).}
SR locates local functional cues, such as an object and its surrounding
regions, by scoring patches in the selected frames against the query words.
Let $p_{ij}$ denote patch $j$ in selected frame $i$, and $W=\{w_m\}_{m=1}^{M}$ the valid text tokens.
We retain $L$ patches per selected frame:
\begin{equation}
    \begin{aligned}
    e_{ij}&=\max_m\operatorname{cos}(p_{ij},w_m),\\
    \mathcal J_{q,i}&=\operatorname{TopL}_{j}(e_{ij}).
    \end{aligned}
\end{equation}
The retained patches form $\widetilde P_q=\{p_{ij}:i\in\mathcal I_q,\ j\in\mathcal J_{q,i}\}$.
Multi-head cross-attention uses these patches as queries and $W$ as keys and values:
\begin{equation}
    \begin{aligned}
    Z_q&=\operatorname{CrossAttn}(\widetilde P_q,W,W),\\
    V^p&=\frac{1}{KL}
    \sum_{i\in\mathcal I_q}\sum_{j\in\mathcal J_{q,i}}z^q_{ij},
    \end{aligned}
\end{equation}
where $z^q_{ij}$ is the refined token corresponding to retained patch $p_{ij}$.
Selection identifies relevant spatial support; cross-attention refines its correspondence with the query words.
The pooled representation $V^p$ summarizes this query-related local evidence.

\noindent\textbf{Temporal Variation Modeling (TV).}
Following temporal difference modeling~\cite{fang2021clip2video,wang2021tdn},
TV represents changes in motion states through short- and long-range frame-feature differences. A multilayer perceptron (MLP) encodes differences
between the original ordered frames:
\begin{equation}
    \begin{aligned}
    \Delta_{\delta}F_i&=F_{i+\delta}-F_i,\\
    V^m&=\frac{1}{2T-h-1}
    \sum_{\delta\in\{1,h\}}\sum_{i=1}^{T-\delta}
    \operatorname{MLP}(\Delta_{\delta}F_i).
    \end{aligned}
\end{equation}
Here, $1<h<T$ specifies a longer temporal interval.
Adjacent-frame differences capture short-term variation, while longer intervals expose gradually accumulating changes.
MLP encoding and averaging over all valid differences produce $V^m$, complementing appearance evidence independently of query-guided selection.
The denominator counts the $T-1$ adjacent-frame differences and $T-h$
longer-range differences. We compute both sets before TS, preserving the
original temporal neighborhoods when the query selects nonadjacent frames.

\subsection{Complementary Fine-Grained Cross-Modal Alignment}
\label{sec:method_alignment}
We compute correspondence at sentence, frame, word, and patch levels.
CFI aggregates matches on query-refined features, and OTA jointly allocates
word--patch correspondence weights on query-independent visual features.

\noindent\textbf{Coarse-to-Fine Interaction (CFI).}
Following coarse-to-fine alignment~\cite{wang2023ucoa}, CFI evaluates
sentence--frame, word--frame, and word--patch correspondence using $(s,F_q)$, $(W,F_q)$, and $(W,Z_q)$, respectively.
For two feature sets $U=\{u_i\}$ and $V=\{v_j\}$, bidirectional aggregation is defined as
\begin{equation}
    \begin{aligned}
    \mathcal B(U,V)=\frac12\Big[
    &\sum_i\rho_i\max_j\operatorname{cos}(u_i,v_j)\\
    +&\sum_j\kappa_j\max_i\operatorname{cos}(u_i,v_j)
    \Big],
    \end{aligned}
\end{equation}
where $\rho_i$ and $\kappa_j$ are normalized weights predicted from the respective features through MLPs followed by softmax.
Each direction aggregates the strongest correspondence of its input units.
The two sums find visual matches for text units and text matches for visual
units, respectively, at each granularity.
The three granularity scores are fused as
\begin{equation}
    \begin{aligned}
    H(t,v)={}&\lambda_{\mathrm{sf}}\mathcal B(\{s\},F_q)\\
    &+\lambda_{\mathrm{wf}}\mathcal B(W,F_q)
    +\lambda_{\mathrm{wp}}\mathcal B(W,Z_q),
    \end{aligned}
\end{equation}
where the nonnegative, learnable weights sum to one.
CFI thus checks overall semantic agreement and the frame-level and local support for individual words.

\noindent\textbf{Optimal-Transport Alignment (OTA).}
CFI aggregates best matches independently, without constraining their joint allocation.
OTA instead establishes a globally coupled correspondence on patch features $Z=\{z_j\}_{j=1}^{N}$ retained from all sampled frames before TS and SR.
Inspired by the marginal-constrained fine-grained alignment in QC-Align~\cite{li2026qcalign}, we predict nonuniform matching budgets for both modalities:
\begin{equation}
    \begin{aligned}
    A_{mj}&=\operatorname{cos}(w_m,z_j),\\
    a&=\operatorname{softmax}(g_t(W)),\quad
    b=\operatorname{softmax}(g_v(Z)),
    \end{aligned}
\end{equation}
where $g_t$ and $g_v$ are independent MLP heads producing one scalar per token.
The distributions $a$ and $b$ specify the total correspondence mass assigned to each word and patch.
The nonuniform marginals assign different matching budgets to words and
patches; all pairwise weights are solved jointly under these budgets.
We solve
\begin{equation}
    \begin{aligned}
    \Pi^*&=\argmax_{\Pi\in\mathcal U(a,b)}
    \langle\Pi,A\rangle+\varepsilon\mathcal H(\Pi),\\
    \mathcal U(a,b)&=
    \{\Pi\geq0:\Pi\mathbf1=a,\ \Pi^\top\mathbf1=b\},
    \end{aligned}
\end{equation}
where $\mathcal H(\Pi)=-\sum_{m,j}\Pi_{mj}\log\Pi_{mj}$.
Sinkhorn iterations~\cite{cuturi2013sinkhorn} alternately update row and column scaling vectors $\mathbf u$ and $\mathbf v$ to approximate these marginal constraints.
The resulting transport plan yields
\begin{equation}
    O(t,v)=\langle\Pi^*,A\rangle.
\end{equation}
OTA thus scores word--patch matches under shared marginal constraints,
using visual features that have not incorporated the query.

For overall matching, we form the text summary $\bar w=\sum_m a_mw_m$ and combine the three evidence representations:
\begin{equation}
    \begin{aligned}
    C(t,v)={}&\mu_g\operatorname{cos}(s,V^g)
    +\mu_p\operatorname{cos}(\bar w,V^p)\\
    &+\mu_m\operatorname{cos}(\bar w,V^m),\\
    S(t,v)={}&C(t,v)+\alpha_{\mathrm{CFI}}H(t,v)
    +\alpha_{\mathrm{OTA}}O(t,v),
    \end{aligned}
\end{equation}
where $\mu_g,\mu_p,\mu_m$ are normalized learnable weights, and the two $\alpha$ coefficients control the fine-grained contributions.
Bidirectional retrieval supervision is applied to $S$ and $C$, treating captions paired with the same video as positives.

\section{Experiments}
\label{sec:experiments}

\begin{table*}[t]
    \centering
    \caption{Bidirectional retrieval on the \airvt test split (\%).
    Directional mR averages R@1, R@5, and R@10; Avg.\ mR averages all
    six recalls. Bold and underline indicate the best and second-best
    scores in each column, respectively.}
    \label{tab:main_results}
    \setlength{\tabcolsep}{4pt}
    \renewcommand{\arraystretch}{1.18}
    \begin{tabular}{llrrrrrrrrr}
        \toprule
        & & \multicolumn{4}{c}{Text-to-Video} & \multicolumn{4}{c}{Video-to-Text} & \\
        \cmidrule(lr){3-6}\cmidrule(lr){7-10}
        Method & Venue & R@1 & R@5 & R@10 & mR & R@1 & R@5 & R@10 & mR & Avg.\ mR \\
        \midrule
        CLIP4Clip~\cite{luo2021clip4clip} & 22'Neurocomputing & 29.20 & 58.07 & 69.68 & 52.31 & \underline{44.93} & 74.28 & 85.09 & 68.10 & 60.21 \\
        BiMa~\cite{le2025bima} & 25'ACM MM & 27.43 & 57.68 & 70.36 & 51.82 & 44.86 & 76.16 & \underline{86.03} & \underline{69.02} & 60.42 \\
        Hybrid-Tower~\cite{lan2025hybridtower} & 25'ICCV & 28.36 & 56.62 & 68.40 & 51.13 & 36.53 & 66.69 & 79.52 & 60.91 & 56.02 \\
        SAVE~\cite{zhao2026save} & 26'CVPR & \underline{29.98} & \underline{59.05} & \underline{71.06} & \underline{53.36} & 44.86 & 74.55 & 84.75 & 68.05 & \underline{60.71} \\
        CoreCaption~\cite{jang2026corecaption} & 26'WACV & 28.44 & 57.59 & 70.05 & 52.02 & 44.06 & \underline{76.83} & \underline{86.03} & 68.97 & 60.50 \\
        MSAM~\cite{huang2026msam} & 26'IEEE TCSVT & 28.84 & 57.36 & 69.47 & 51.89 & 44.80 & 73.20 & 84.15 & 67.38 & 59.64 \\
        \midrule
        AirFocus-CST & Ours & \textbf{32.36} & \textbf{62.16} & \textbf{73.75} & \textbf{56.09} & \textbf{49.36} & \textbf{78.11} & \textbf{86.30} & \textbf{71.26} & \textbf{63.67} \\
        \bottomrule
    \end{tabular}
\end{table*}

\begin{table*}[t]
    \centering
    \caption{Bidirectional retrieval on CapERA (\%). Bold and underline denote the best and second-best scores in each column.}
    \label{tab:capera_results}
    \setlength{\tabcolsep}{4pt}
    \renewcommand{\arraystretch}{1.18}
    \begin{tabular}{llrrrrrrrrr}
        \toprule
        & & \multicolumn{4}{c}{Text-to-Video} & \multicolumn{4}{c}{Video-to-Text} & \\
        \cmidrule(lr){3-6}\cmidrule(lr){7-10}
        Method & Venue & R@1 & R@5 & R@10 & mR & R@1 & R@5 & R@10 & mR & Avg.\ mR \\
        \midrule
        CLIP4Clip~\cite{luo2021clip4clip} & 22'Neurocomputing & \underline{7.29} & 22.27 & 35.07 & \underline{21.54} & \textbf{11.72} & \underline{25.09} & \underline{33.57} & \underline{23.46} & \underline{22.50} \\
        BiMa~\cite{le2025bima} & 25'ACM MM & 6.08 & \underline{22.37} & \underline{35.36} & 21.27 & 8.91 & 23.58 & 32.78 & 21.76 & 21.51 \\
        Hybrid-Tower~\cite{lan2025hybridtower} & 25'ICCV & 6.04 & 19.58 & 31.30 & 18.97 & 7.98 & 19.19 & 28.04 & 18.40 & 18.69 \\
        SAVE~\cite{zhao2026save} & 26'CVPR & 6.00 & 20.65 & 33.21 & 19.95 & 8.27 & 19.91 & 29.04 & 19.08 & 19.51 \\
        CoreCaption~\cite{jang2026corecaption} & 26'WACV & 5.03 & 19.31 & 31.23 & 18.52 & 8.05 & 18.98 & 27.53 & 18.19 & 18.36 \\
        MSAM~\cite{huang2026msam} & 26'IEEE TCSVT & 5.84 & 20.30 & 31.95 & 19.36 & 8.20 & 20.85 & 29.48 & 19.51 & 19.43 \\
        \midrule
        AirFocus-CST & Ours & \textbf{8.17} & \textbf{26.53} & \textbf{40.76} & \textbf{25.15} & \underline{9.49} & \textbf{25.16} & \textbf{36.38} & \textbf{23.68} & \textbf{24.41} \\
        \bottomrule
    \end{tabular}
\end{table*}

\subsection{Experimental Settings}
\label{sec:exp_setup}
We implement all experiments in PyTorch 2.8.0 (CUDA 12.8) on a single NVIDIA A100-SXM4 GPU (40 GB).
We uniformly sample 16 frames over the duration of each video and apply standard CLIP
preprocessing at a resolution of $224\times224$. The frozen CLIP ViT-B/32 visual
backbone produces 49 patch tokens per frame. Preprocessing resizes the shorter
image edge to 224 pixels, applies a center crop, and uses CLIP normalization.
The temporal encoder in the TV branch uses two Transformer layers with pre-layer normalization (pre-LN) and a hidden dimension of 512.
We train \airfocus
for 20 epochs using AdamW with an initial learning rate and weight decay both set to
$10^{-4}$, and cosine annealing without warmup. We clip gradient norms at 1.0.
On \airvt, each training batch contains eight videos and all
ten descriptions per video, for a total of 80 video--text pairs. We select the
checkpoint on the validation split and evaluate it on the held-out \airvt test split.
Evaluation follows Sec.~\ref{sec:dataset_protocol}, reporting R@1, R@5,
R@10, directional mR, and bidirectional Avg.\ mR.

\subsection{Results Comparison}
\label{sec:exp_main}
We compare \airfocus with six methods, including general video--text retrieval models and the drone-specific MSAM. On \airvt (Tab.~\ref{tab:main_results}), \airfocus ranks first at R@1, R@5, and R@10 in both directions and achieves an Avg.\ mR of 63.67\%, exceeding SAVE by 2.96 percentage points. In T2V retrieval, its R@1 and mR reach 32.36\% and 56.09\%, improving over SAVE by 2.38 and 2.73 points. In V2T retrieval, R@1 and mR reach 49.36\% and 71.26\%, exceeding the strongest respective baselines, CLIP4Clip and BiMa, by 4.43 and 2.24 points. V2T gains are concentrated at rank one: the R@1 improvement is 4.43 points, compared with 0.27 points at R@10.

To evaluate the method on a second aerial benchmark, we conduct comparisons on CapERA~\cite{bashmal2023capera}. As shown in Tab.~\ref{tab:capera_results}, \airfocus achieves an Avg.\ mR of 24.41\%, outperforming CLIP4Clip by 1.91 points. It leads at all three T2V recall cutoffs, raising directional mR from 21.54\% to 25.15\%. In V2T retrieval, it achieves the highest mR (23.68\%) and R@10 (36.38\%); CLIP4Clip ranks first at R@1 (11.72\%, versus 9.49\%). The larger T2V improvement identifies text-conditioned video ranking as the main source of the overall gain on CapERA.

\subsection{Effectiveness Analysis}
\label{sec:exp_effectiveness}

\begin{figure*}[t]
    \centering
    \includegraphics[width=\textwidth]{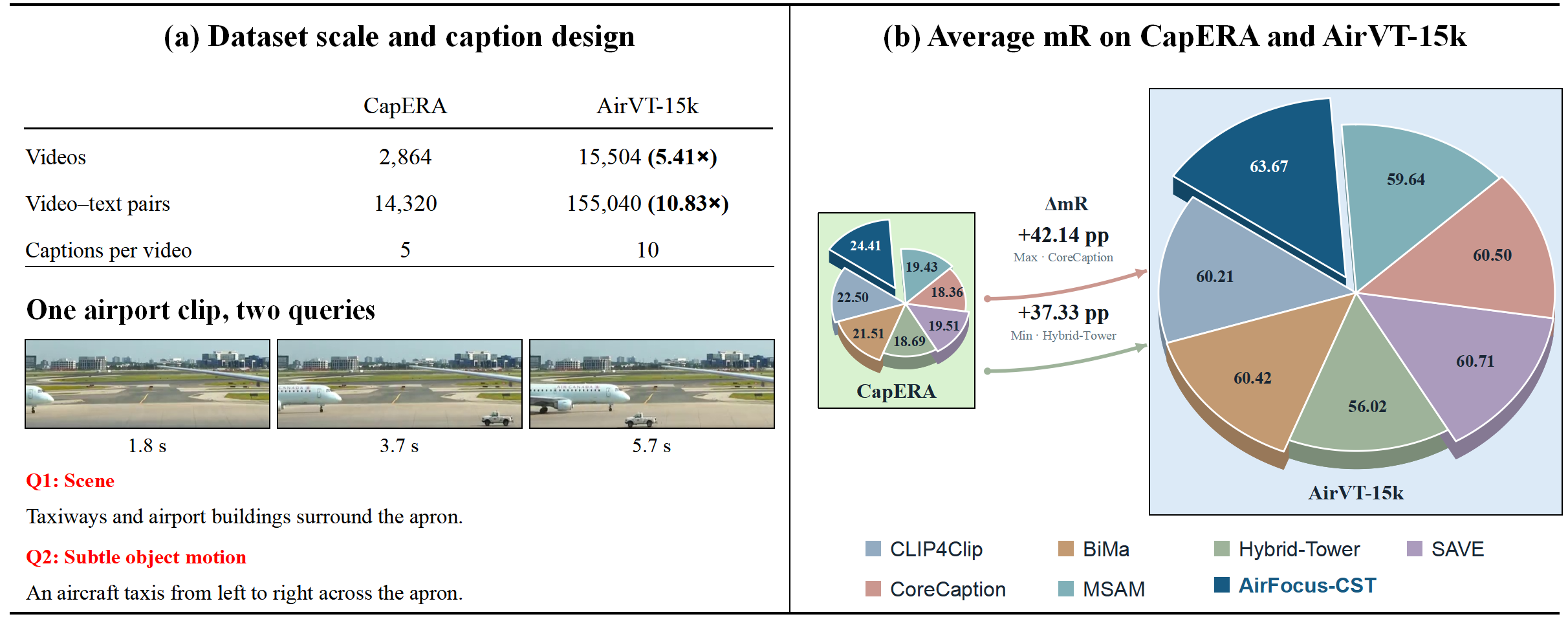}
    \caption{Effectiveness analysis. (a) Dataset scale, annotation
    design, and complementary queries from the same AirVT-15k clip.
    (b) Average mR of seven methods on CapERA and AirVT-15k.
    Gap denotes the difference between datasets for the same method;
    Max and Min identify the largest and smallest such differences.}
    \label{fig:effectiveness}
\end{figure*}
\paragraph{Dataset effectiveness.}
Figure~\ref{fig:effectiveness}(a) compares dataset scale and caption content. \airvt contains 15,504 videos and 155,040 captions, compared with 2,864 videos and 14,320 captions in CapERA~\cite{bashmal2023capera}, giving 5.41 times as many videos and 10.83 times as many captions. CapERA provides five captions per video; \airvt provides ten. The airport example shows the distinction in content: one caption describes the apron and nearby buildings, and another follows an aircraft taxiing across the apron. The two queries refer to the same clip but require scene context and object motion, respectively. Figure~\ref{fig:effectiveness}(b) shows higher overall retrieval scores on \airvt for every method, including SAVE and CoreCaption. This pattern spans frame aggregation in CLIP4Clip, semantic enrichment in SAVE and CoreCaption, and drone-specific matching in MSAM.

\paragraph{Method effectiveness.}
\airfocus occupies the top position within both dataset groups in Fig.~\ref{fig:effectiveness}(b). It also ranks first on most individual metrics in both retrieval directions (Tab.~\ref{tab:main_results}). The ablations below measure the contributions of each evidence source and alignment module.

\subsection{Ablation Study}
\label{sec:exp_ablation}

\begin{table*}[t]
    \centering
    \caption{Component ablations on \airvt.
    TS: query-guided temporal selection; SR: query-guided spatial
    refinement; TV: temporal variation modeling.
    CFI and OTA denote the two alignment components.
    Checkmarks indicate enabled components; global scene context is
    retained in every configuration.}
    \label{tab:ablation}
    \setlength{\tabcolsep}{5pt}
    \begin{tabular}{clccccc rrr}
        \toprule
        & & \multicolumn{3}{c}{Multi-evidence modeling}
        & \multicolumn{2}{c}{Alignment}
        & \multicolumn{2}{c}{R@1} & \\
        \cmidrule(lr){3-5}\cmidrule(lr){6-7}\cmidrule(lr){8-9}
        ID & Configuration & TS & SR & TV & CFI & OTA
        & T2V & V2T & Avg.\ mR \\
        \midrule
        A & Global-only baseline & -- & -- & -- & -- & --
          & 29.42 & 44.86 & 60.22 \\
        B & Multi-evidence only & \checkmark & \checkmark & \checkmark & -- & --
          & 30.71 & 46.94 & 61.77 \\
        C & Multi-evidence + CFI & \checkmark & \checkmark & \checkmark & \checkmark & --
          & 31.58 & 48.15 & 63.01 \\
        D & Multi-evidence + OTA & \checkmark & \checkmark & \checkmark & -- & \checkmark
          & 31.27 & 47.62 & 62.63 \\
        \midrule
        E & w/o temporal selection & -- & \checkmark & \checkmark & \checkmark & \checkmark
          & 31.62 & 48.02 & 63.03 \\
        F & w/o spatial refinement & \checkmark & -- & \checkmark & \checkmark & \checkmark
          & 31.44 & 48.09 & 62.87 \\
        G & w/o temporal variation & \checkmark & \checkmark & -- & \checkmark & \checkmark
          & 31.09 & 47.55 & 62.39 \\
        \midrule
        H & \textbf{AirFocus-CST (full)} & \checkmark & \checkmark & \checkmark & \checkmark & \checkmark
          & \textbf{32.36} & \textbf{49.36} & \textbf{63.67} \\
        \bottomrule
    \end{tabular}
\end{table*}
Table~\ref{tab:ablation} separates evidence modeling from cross-modal alignment. Adding TS, SR, and TV to the global-only baseline raises Avg.\ mR by 1.55 percentage points (A vs.\ B). CFI and OTA then add 1.24 and 0.86 points individually (B vs.\ C/D). Using both gives a further 0.66 points over CFI alone and 1.04 points over OTA alone (C/D vs.\ H). Each alignment module therefore contributes beyond the other: query-refined local interactions and query-independent transport provide matching scores.

Removing TS, SR, or TV reduces Avg.\ mR by 0.64, 0.80, and 1.28 points, respectively (E/F/G vs.\ H), and lowers R@1 in both directions. TV has the largest effect among the three evidence modules even though configuration G retains both alignment modules. The comparison isolates the benefit of temporal feature changes when query-guided frame selection, spatial refinement, and both alignment modules are enabled.

The full model improves both retrieval directions over the global-only baseline, with gains of 2.94 points in T2V R@1 and 4.50 points in V2T R@1 (A vs.\ H). Avg.\ mR rises by 3.45 points in the same comparison. The complete configuration improves first-ranked matches in both retrieval directions as well as mean recall across cutoffs. With CFI already enabled, adding OTA further improves T2V and V2T R@1 by 0.78 and 1.21 points, respectively (C vs.\ H), supporting joint word--patch allocation beyond independent best-match aggregation.

\section{Conclusion}
\label{sec:conclusion}
We introduced \airvt, a UAV video--text benchmark whose ten complementary captions per video describe both the surrounding scene and changes in object activity. Each clip therefore supports queries about either its scene layout or its object activity. Our model, \airfocus, selects query-relevant frames and patches, encodes temporal feature differences, and combines CFI with OTA for cross-modal matching. It achieves the highest Avg.\ mR on both \airvt and CapERA. Ablations show that removing temporal variation causes the largest drop among the three evidence modules and that each alignment module adds gains beyond the other.

\section*{LLM Usage Statement}
We used ChatGPT-6 for language polishing and grammar checking, and to generate initial English--Chinese caption drafts for AirVT-15k. These drafts were subsequently checked and corrected by the authors. All AI-assisted revisions were carefully reviewed and approved by the authors, who remain fully responsible for the accuracy and integrity of the manuscript.

{
    \small
    \bibliographystyle{ieeenat_fullname}
    \bibliography{main}
}

\end{document}